\documentclass[conference]{IEEEtran}
\IEEEoverridecommandlockouts
\usepackage{cite}
\usepackage{amsmath,amssymb,amsfonts}
\usepackage{algorithmic}
\usepackage{graphicx}
\usepackage{textcomp}
\usepackage{xcolor}
\usepackage{glossaries}
\usepackage{multirow}

\newacronym{AI}{AI}{Artificial Intelligence}
\newacronym{DFS}{DFS}{Depth-First Search}
\newacronym{RRT}{RRT}{Rapidly-Exploring Random Tree}
\newacronym{PID}{PID}{Proportional-Integral-Derivative}
\newacronym{MARL}{MARL}{Multi-Agent Reinforcement Learning}
\newacronym{A3C}{A3C}{Advantage
Actor Critic}
\newacronym{ANQ}{ANQ}{Asynchronous $n$-step Q-Learning}
\def\BibTeX{{\rm B\kern-.05em{\sc i\kern-.025em b}\kern-.08em
    T\kern-.1667em\lower.7ex\hbox{E}\kern-.125emX}}
\begin{document}

\title{A Game AI Competition to foster Collaborative AI research and development
}


\author{\IEEEauthorblockN{Ana Salta\IEEEauthorrefmark{1},
Rui Prada\IEEEauthorrefmark{2} and Francisco S. Melo\IEEEauthorrefmark{3}}
\IEEEauthorblockA{INESC-ID \\
Instituto Superior Técnico, Universidade de Lisboa\\
Porto Salvo, Portugal \\
Email: \IEEEauthorrefmark{1}anasalta@tecnico.ulisboa.pt,
\IEEEauthorrefmark{2}rui.prada@tecnico.ulisboa.pt,
\IEEEauthorrefmark{3}fmelo@inesc-id.pt}}

\maketitle

\begin{abstract}
Game AI competitions are important to foster research and development on Game AI and AI in general. These competitions supply different challenging problems that can be translated into other contexts, virtual or real. They provide frameworks and tools to facilitate the research on their core topics and provide means for comparing and sharing results. A competition is also a way to motivate new researchers to study these challenges. In this document, we present the Geometry Friends Game AI Competition. Geometry Friends is a two-player cooperative physics-based puzzle platformer computer game. The concept of the game is simple, though its solving has proven to be difficult. While the main and apparent focus of the game is cooperation, it also relies on other AI related problems such as planning, plan execution, and motion control, all connected to situational awareness. All of these must be solved in real-time. In this paper we discuss the competition and the challenges it brings, and present an overview of the current solutions.


\end{abstract}


\section{Introduction}\label{sec:Intro}
Game \gls{AI} has been around for several decades. Many computer games make use of \gls{AI}, whether to serve as the players' opponents or allies, to create real-time challenges by altering the environment or even generate entire new levels. Although the main value of computer games is entertainment, they can also be used to improve other aspects of the real or virtual world. For example, some games have educational value as well. Games based on vehicle simulators, for instance, can teach and train piloting skills that players can later apply in real-life. A similar approach could be used with artificial autonomous agents. An agent that can pilot a simulated car in a game, so it can be used as a rival to the player, during a race, could later be used as a base for self driving cars. It all depends on the level of detail of said simulation, but in the end, this kind of development is done with several steps of incremented complexity.

There is a wide variety of skills that games can teach. Multiplayer games, for instance, can teach social skills. In those games, players may find themselves competing against other players or cooperating with them. While the competition may focus more on the individual skills, cooperation allows players to develop skills as team members. 

Competitions of Game \gls{AI}, similarly to other AI fields \cite{kitano:robocup}, have been used frequently and successfully to foster the development of new and innovative research. Competitions motivate many researchers to work on difficult problems and provide, at the same time, a framework for common ground to contextualise and compare research. Also, competitions may be used to support teaching of Game \gls{AI} (and \gls{AI} in general) as they present very concrete problems for students to address and, typically, share openly the submissions and results.

In this paper, we present the Geometry Friends Game \gls{AI} Competition~\cite{GF} that runs around a physics based collaborative puzzle-platformer game (Geometry Friends). Because of its cooperative nature, the game promotes interesting \gls{AI} challenges that are not often seen in game \gls{AI} competitions. It provides challenges of collaboration and team AI at different levels. In the paper, we discuss some of the results obtained so far and the implications and potential importance of the competition for game \gls{AI} research. We start by explaining the Geometry Friends game in section~\ref{sec:Game}, after some discussion of related work on competitions of game AI, in section~\ref{sec:RW}. In section~\ref{sec:problems} we enumerate the \gls{AI} problems that can be explored with the game. We then present the competition and framework in the respective sections~\ref{sec:Competition} and \ref{sec:FrameWork}. Afterwards, we describe briefly the solutions developed so far, in section~\ref{sec:solutions}, most of which competed in previous competitions. Finally, we draw some conclusions in section~\ref{sec:conclusions} and discuss future work.

\section{Game AI Competitions}\label{sec:RW}
Game AI competitions have been taking place for several decades. 

For example, the Computer Olympiad ran by ICGA\footnote{www.icga.org}, started in 1989.  It challenges participants to submit agents, or programs, that can play one or more games, mostly board and card games. 

The Battlecode\footnote{www.battlecode.org} is described as ``MIT's longest-running programming competition''. It presents a turn-based strategic game where two teams of agents compete against each other having to manage resources and use combat tactics. Each year, different challenges are added to the game.

There are some competitions that include collaboration as a challenge for the Game AI and some other competitions address Physics-Based Simulation Games~\cite{PBSG}. Both these topics are relevant in the Geometry Friends competition, but we will focus our discussion here in the collaboration aspects that are more central and unique in the competition.

The General Video Game AI Competition\cite{GVGAI} challenges the participants to develop agents that can play any game, without any prior knowledge of it. The agents are tested with a game that they do not know the rules of. It presents to participants many different kinds of games, and more recently, included collaborative games as well.

The Malmo Collaborative AI Challenge\footnote{https://www.microsoft.com/en-us/research/academic-program/collaborative-ai-challenge/}, presents a mini-game based on the ``stag-hunt'' paradigm, implemented with Project Malmo\cite{Malmo}, an experimentation and research platform for AI, built on top of the Minecraft game. The challenge was designed to encourage research related to various problems in Collaborative AI. It invites participants to develop collaborative AI solutions to achieve high scores across a range of partners. In the game presented players need to work together to achieve a common goal, as is the case of Geometry Friends.

Furthermore, the Hanabi competition~\cite{Hanabi} introduces the collaborative turn-based card game with two possible tracks. In one of the tracks the agents follow the same strategy, while in the other they are not aware who they are paired with. The game focus on the ``why'' of the other player's intentions, so a model of the other should be created. It is also about understanding how the other players will understand the instructions given to them. 

The FruitPunch AI-esports competition\footnote{www.fruitpunch.ai/competition/} currently has a multiplayer human-agent cooperative game where the challenges are smart navigation, decision making and risk management. In this game, named ``Isaac's Labyrinth'', there are multiple teams of two players: one agent and one of their developers. The objective of the game is to navigate through a dynamic maze to catch some fruits. While the agent performs the navigation, the human player can place or reveal traps. This competition focus on approaches that use reinforcement learning. 

These are just some examples of Game \gls{AI} competitions. Some of these share similar problems with the ones in Geometry Friends, but none deal with them in the same way, nor all at the same time. Card and board games usually focus on developing or improving search algorithms, or other generalized strategies that may rely on patterns. In Geometry Friends, this is also important for planning and control. Some of the games presented in these competitions are turn-based, and although they might have time limits to perform their actions at each turn, there is no real-time issue being explored. ``Isaac's Labyrint'' has also the component of adversaries, and focus on human-agent cooperation, which may be a future possibility for the Geometry Friends competition since the game allows human players to play with artificial agents. 
Hanabi's focus on the understanding of the partner's intentions using implicit communication that can be explored on Geometry Friends as well.

\section{Geometry Friends Game}\label{sec:Game}
\begin{figure}[h]
\centerline{\includegraphics[width=\columnwidth]{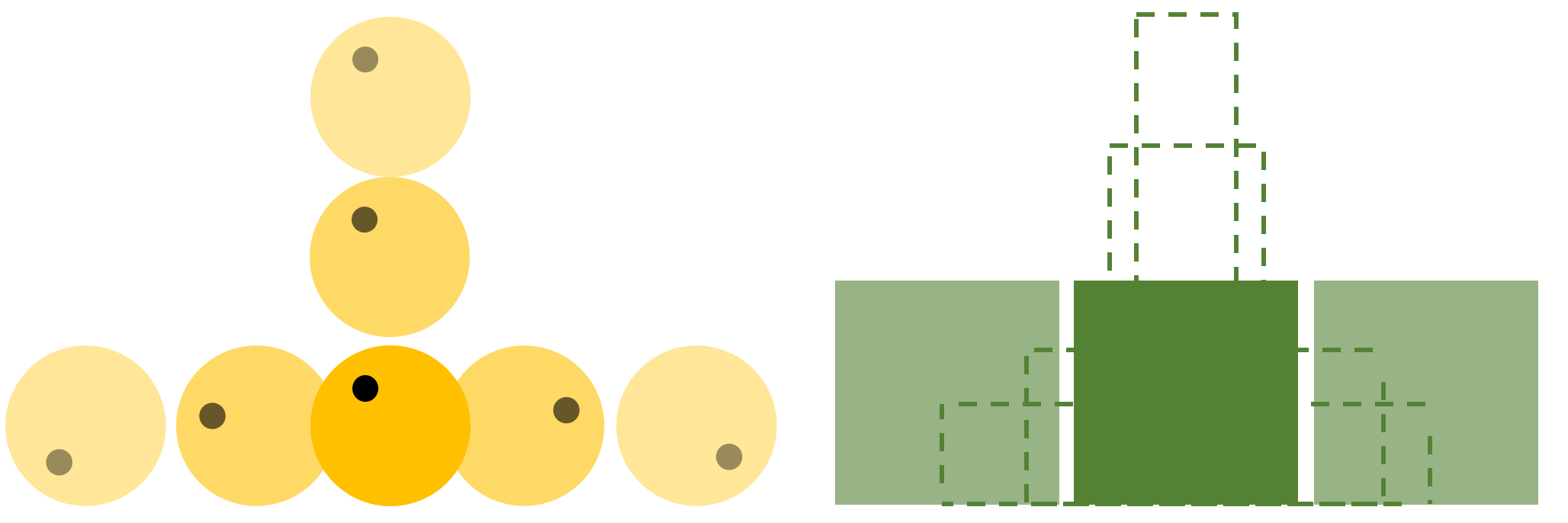}}
\caption{Circle and rectangle characters and their possible actions} \label{fig:characters}
\end{figure}

\begin{figure}[h]
\centering
\begin{minipage}{.5\columnwidth}
\centering
\includegraphics[width=\columnwidth]{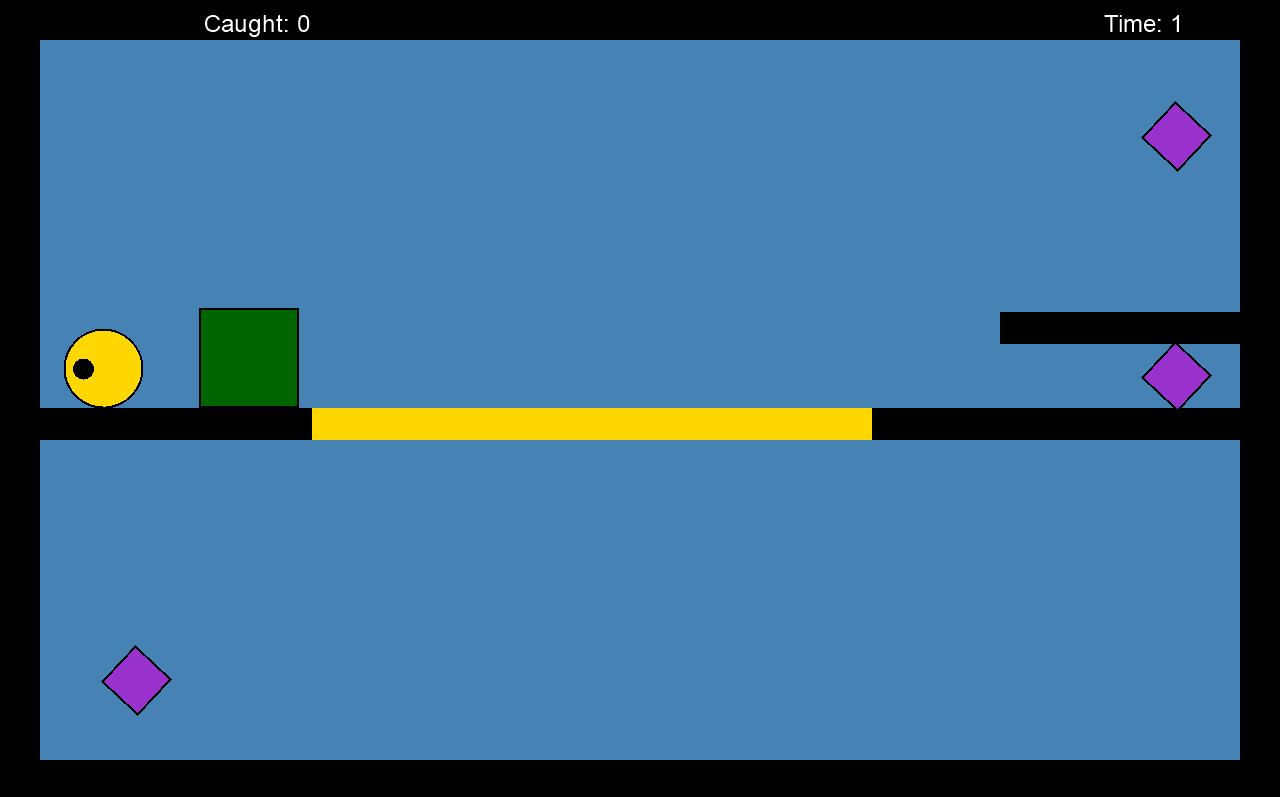}
\end{minipage}%
\begin{minipage}{.5\columnwidth}
\centering
\includegraphics[width=\columnwidth]{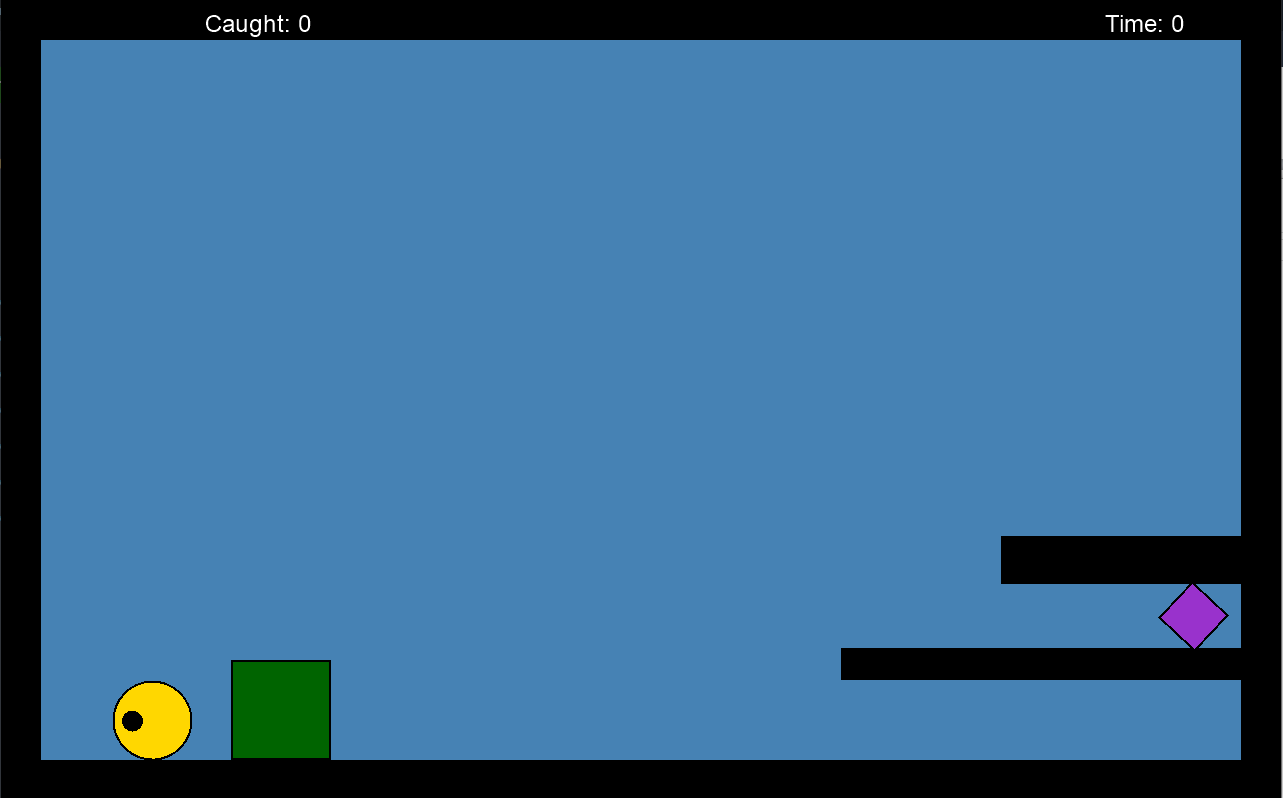}
\end{minipage}
\caption{Sample cooperative levels from the 2017 competition} \label{fig:cooplevel01}
\end{figure}

Geometry Friends is a collaborative physics-based puzzle and platform game. It has two characters, a circle and a rectangle, each with different possible actions, as shown in Fig.~\ref{fig:characters}: jump and roll, for the circle, and morph and slide, for the rectangle.
The game is composed by a series of levels in a physics-based 2D world. Each level is defined by a set of platforms and collectibles (diamonds) that the characters need to collect. The kind of challenges presented vary according to the shape and position of the platforms, the position of the diamonds and the initial position of the two characters. For example, catching some diamonds might require motion coordination between the two characters, while in other cases it will be better to split the task and each character catch part of the diamonds by themselves. The platforms can be of one of three colours. The characters will only collide with platforms that have a different colour than themselves. The rectangle will collide against black and yellow platforms, and ignore the green ones. The circle will collide against black and green platforms, and ignore the yellow ones. Figure~\ref{fig:cooplevel01} shows two examples of cooperative levels. In the level on the left, there are two diamonds that can only be caught by the circle (the highest and the lowest), and one only by the rectangle. Although the circle can reach the highest diamond, it can only do so with the help of the rectangle, since the yellow platform is not considered as a support by the circle. This means the circle needs to ``ride'' the rectangle to reach the other side of the screen.

The game also has single-player levels. Those are similar to the cooperative, but only one of the characters is placed in the game world. The goal is the same, to collect all diamonds. These levels, help testing agents in the role of each of the characters separately.
\section{AI PROBLEMS TO EXPLORE}\label{sec:problems}
As simple as the Geometry Friends game appears to be, it deals with several problems of great interest to \gls{AI} research. The game presents several situations of combined task and motion planning, in a dynamic physics-based world, that agents need to deal in real-time. And, more importantly, it presents collaborative challenges of different perspectives. 

\begin{figure}[h]
\centerline{\includegraphics[width=\columnwidth]{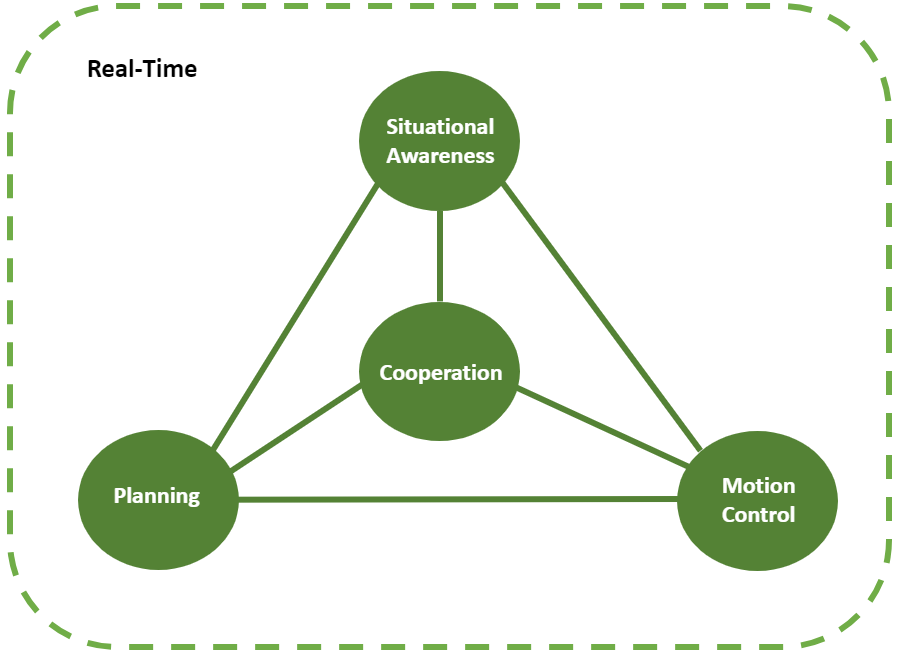}}
\caption{AI problems that can be explored in Geometry Friends} \label{fig:problems}
\end{figure}

The main \gls{AI} problem the competition focuses on is cooperation. But, it includes serious challenges for other subjects as well (see Fig.~\ref{fig:problems}). The game involves planning, motion control, and situational assessment. Hence to develop successful agents for Geometry Friends, all these should be taken into consideration in a balanced way. On top of that, each of the subjects must consider cooperation on its own perspective. For example, to define common plans, to execute actions together and to have in consideration the others' view when assessing the game situations.

\subsection{Cooperation}

As stated, the main objective of the competition is to encourage the participants to develop a team of two agents capable of solving cooperative levels. In this context, we define cooperation as a ``set of actions made by the two agents in order to achieve a common goal''. The decision to cooperate or not is not addressed. We are concerned with the aspects that lead the agents to good cooperation, which is measured through their performance in the game. A collaborative challenge is one that entails that if the agents achieve cooperation their performance will improve. In fact, in Geometry Friends, it is often the case that the level is impossible if good cooperation is not achieved.


There are several important aspects to address in order to achieve good cooperation. As the motivation to cooperate is assumed by the fact that both agents are on the same team and they have no individual rewards on the game, most problems are related to coordination. By coordination we mean that the actions that the two agents take must be properly sequenced and timed as there are dependencies between them.

\subsubsection{Task Division}

The cooperative levels may have problems where to be efficient and effective, the agent must decide the best order to catch a diamond. If one diamond can be caught by both agents, they should be able to decide which of them will do it, and not waste time by having them both trying to do so. They should also know how to divide the tasks in a way no agent needs to wait more than necessary time to catch a diamond in need of joint action (see bellow). In fact, the division of task should identify which parts of the task can be solved by each individual alone and which parts need simultaneous action.

\begin{figure}[h]
\centering
\begin{minipage}{.5\columnwidth}
\centering
\includegraphics[width=\columnwidth]{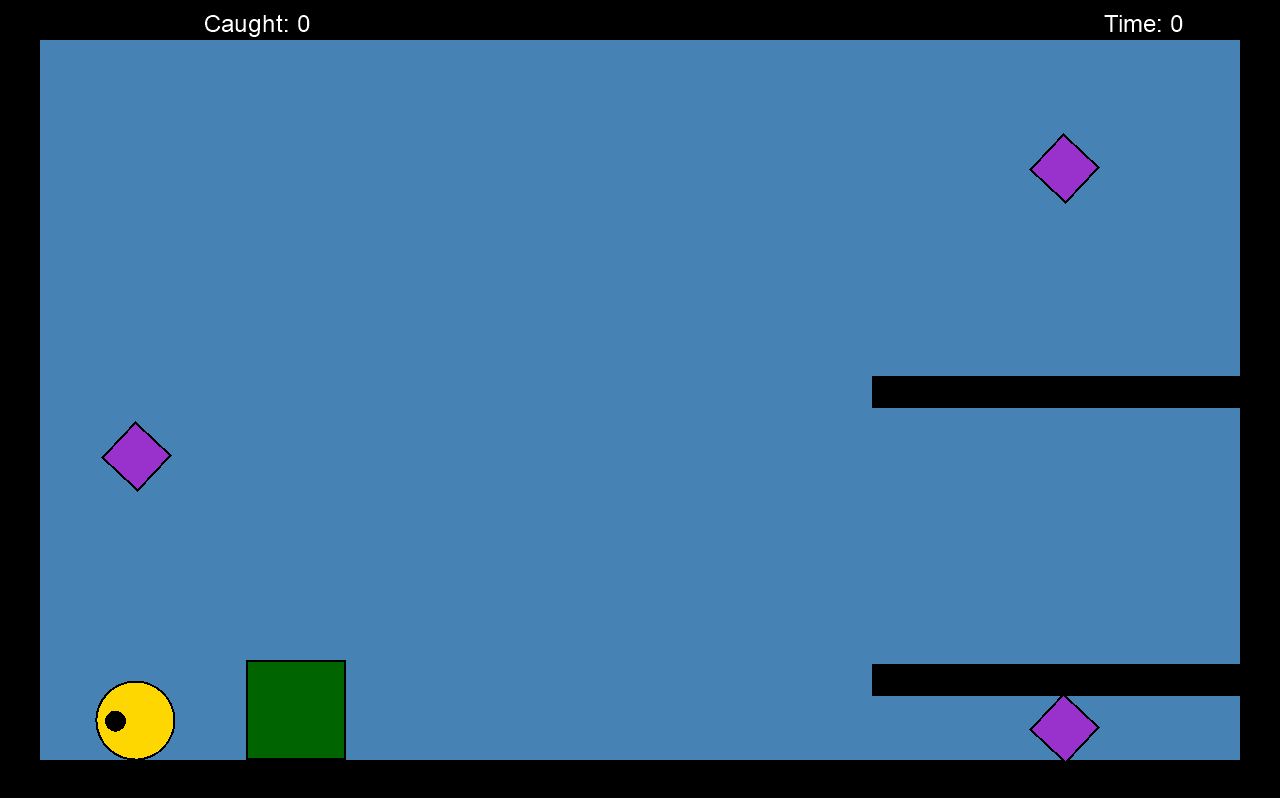}

\end{minipage}%
\begin{minipage}{.5\columnwidth}
\centering
\includegraphics[width=\columnwidth]{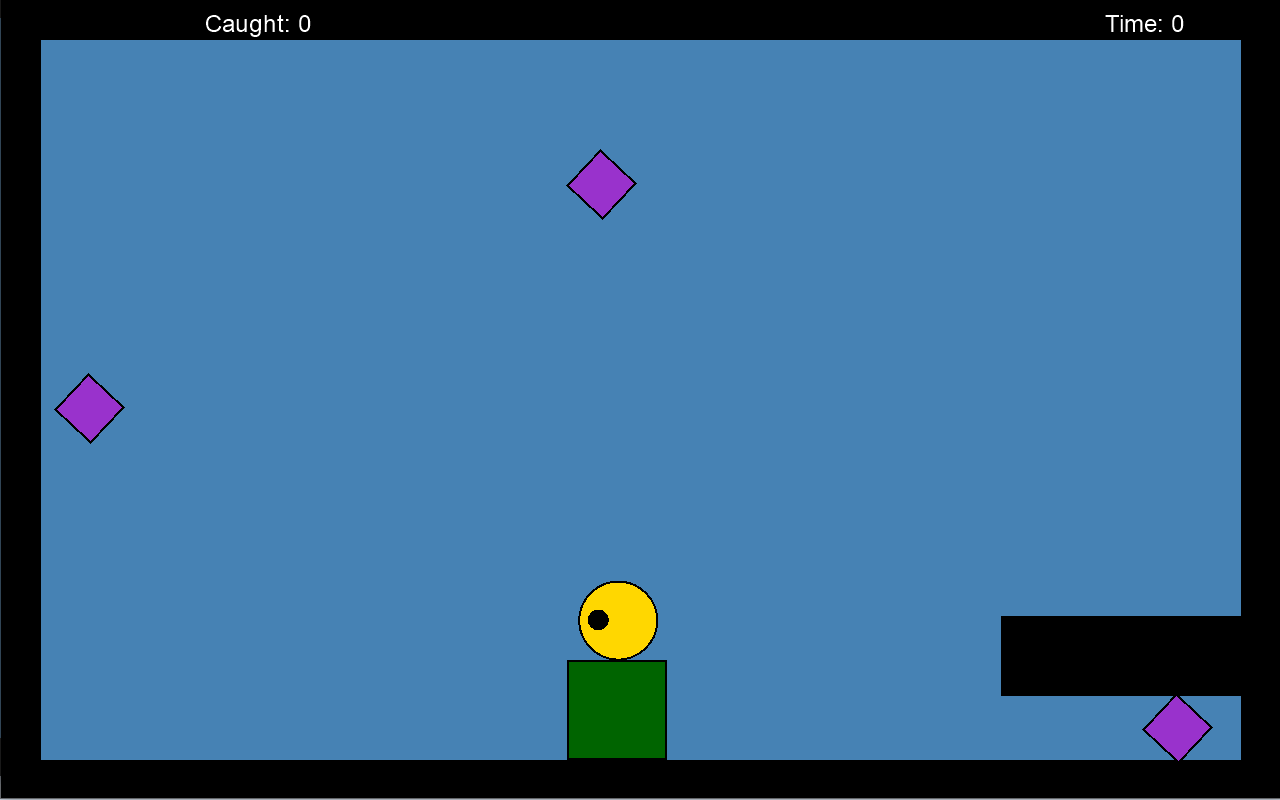}
\end{minipage}
\caption{Cooperation with different task division strategies} \label{fig:cooplevel04}
\end{figure}

Figure~\ref{fig:cooplevel04}, on the left, presents an example of a level that can be solved faster if the agents catch first their respective diamonds (for the circle the one right above, and for the rectangle the one on the right-low corner) and then go for the one that requires cooperation. 

The level shown on the right of the figure presents the opposite situation. The agents start at a position that makes their joint action easy and quick to perform as the first set of the cooperation. Not taking advantage of their initial positions will make it harder to catch the highest diamond and lose time.

\subsubsection{Joint Action}

Many of the cooperative levels present situations that require both characters to perform coordinated actions at the same time. 
For example, as shown before in the Figure~\ref{fig:cooplevel01} (left), the circle may need to ride the rectangle to reach other places in the level, this action needs good timing to make sure the characters move together. In general, the circle may use the rectangle as another (eventually moving) platform to jump on. This is often the case to help the circle to reach higher ground. But, the rectangle may need the help of the rectangle to climb a platform, as shown in Figure~\ref{fig:cooplevel01} (right), or to cross from a platform to another, as it lacks the ability to jump. In the first case the rectangle can slide against the circle to tilt to the right, and in the latter case, the circle needs to collide with the rectangle, to apply additional force to its movement, at the right moment.


\subsubsection{Communication}\label{subsubsec:communication}

Communication is important for cooperation. The Geometry Friends game supports research on either explicit or implicit communication. 

We see explicit communication as the exchange of signals with pre-established meaning. The competition framework allows the agents to send messages that can be used as a channel of explicit communication. On the other hand, agents may attempt of sending messages using the other available actions, or implying intents through the observation of the partner's behavior. For example, the circle may try to catch the rectangle's attention to a certain position, by rolling to the place and jumping nonstop, expecting the rectangle to understand the message it is trying to convey. If the meaning of this sort of action is not previously known to both agents, then it counts as implicit communication. This kind of communication is more challenging and also of great importance for \gls{AI} research, including game \gls{AI}. 

If one of the agents is human, we are in a situation of human-agent interactions. In this case, we can research good ways to communicate using restrictive channels. The agents may be able to quickly transmit useful information to the player either through audio or visual signals, but the other way around may be harder. The players may need to be able to reference certain positions in the level by point, for example, but this may not be enough to make their intentions clear. Reading and learning from the actions of the other players will be important for the agents. Another approach is to use Natural Language. There is an increasing interest in the use of Natural Language as a natural interface for AI systems, and there have been several achievements that make this approach promising.

The Geometry Friends game is an interesting test-bed for communication as it deal with communication from a collaborative perspective. Communication can be used to support mutual understanding, to share different perspectives about the situation awareness, to devise plans together, to convey intentions, to coordinate actions and even for motivation and emotional support.


\subsection{Situational Awareness}

Situational Awareness is about the perception of the environment, but also includes comprehension of the perceived state, which supports the prediction of the future states\cite{Endsley1995}.

The perceptions are the input given by the sensors of the agent. The game world is fully observable, hence the agents can get information about the full state of the level (e.g. the position of all elements). The perception is not a challenge, but the comprehension comes with some challenges. On the one hand, the dynamic part of the characters' movement is not given by the sensors, for example, the agents do not receive the information about the angular velocity of the character or the force that is being applied to generate movement. On the other hand, while it is fairly easy to build a representation of the state of the level, it is not as easy to understand what the agents can do in the level. Agents, should build representations of the level that include information about the action affordances as well, for example, how platforms may be connected and what are the possible jump points for the circle. This should also include the identification of situations that may render the conclusion of the level impossible (e.g. falling into a pit). Note that in the presence of coloured platforms, the areas that are available for each of the characters are typically different.

To assess the situation in this game, it is also important to identify which diamonds need joint action and which do not, and understand the intentions of the partner and what it is doing to be able to coordinate the actions. 

\subsection{Planning}\label{sebsec:planning}

In Geometry Friends agents need to consider futures states for action, in general to come up with a plan for action to solve the level in an efficient way, but also to avoid ending in states that make the level impossible. The game presents opportunities for hierarchical planning as it combines challenges of both task and motion planning. Task planning is more strategic and involves deciding the general movement around the level (e.g. which platforms to take, and in what order) and coordination actions (e.g. where and when the two characters should meet to perform a join action). The motion planning, is more low level, and involves deciding the control actions needed to move around the level correctly.

\begin{figure}[h]
\centering
\begin{minipage}{.5\columnwidth}
\centering
\includegraphics[width=\columnwidth]{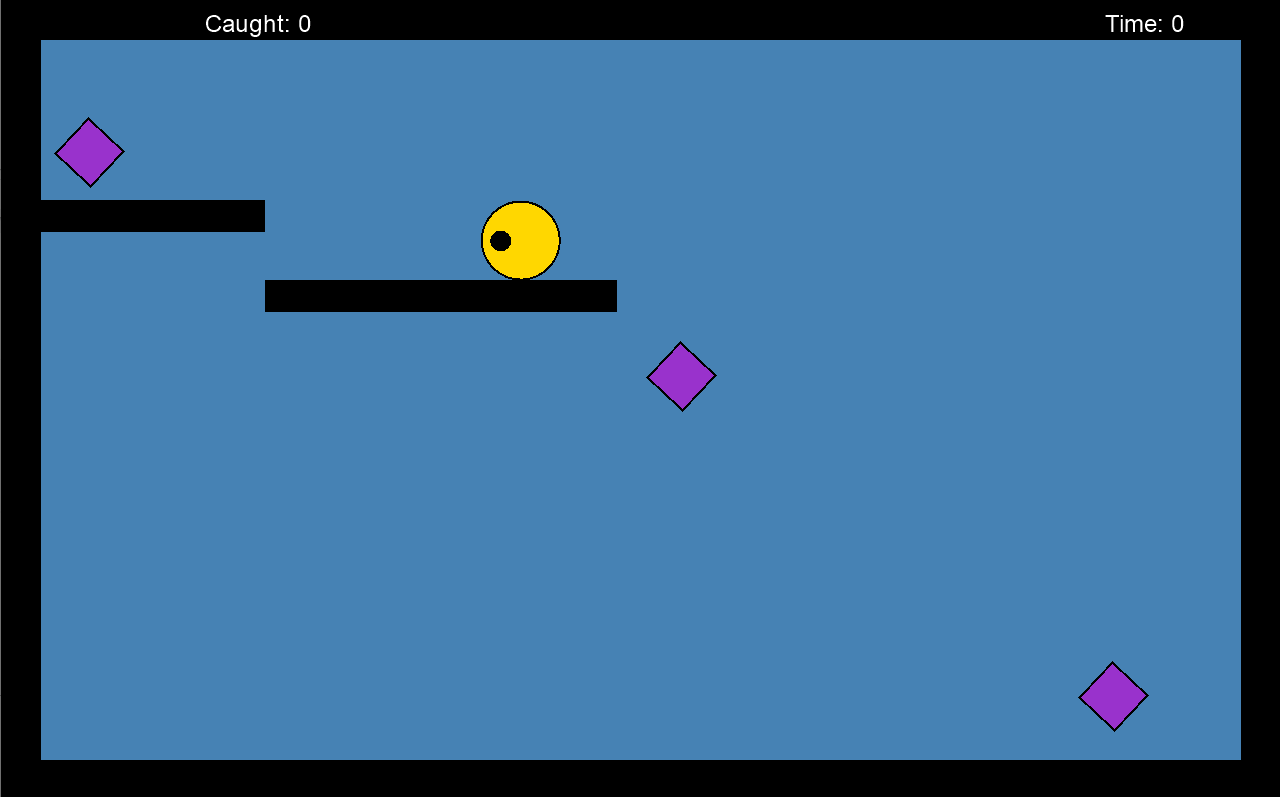}
\end{minipage}%
\begin{minipage}{.5\columnwidth}
\centering
\includegraphics[width=\columnwidth]{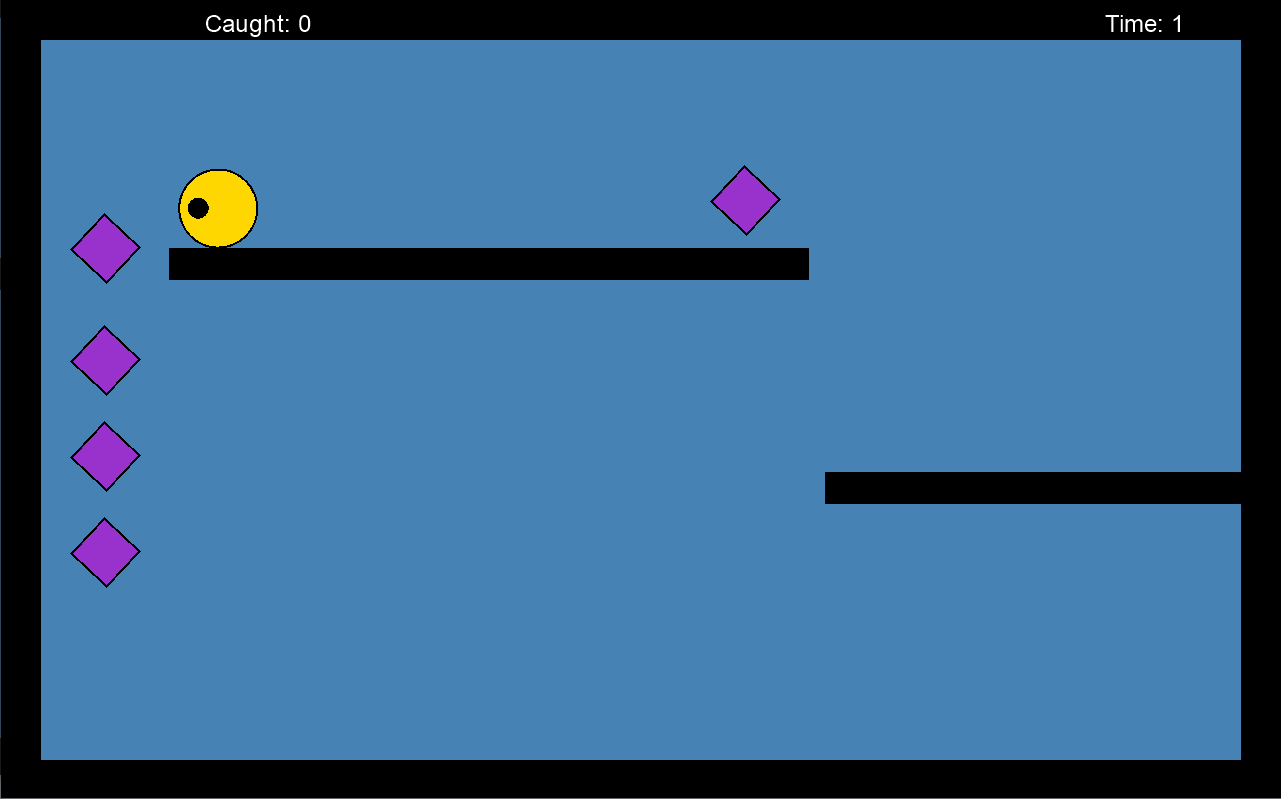}
\end{minipage}
\caption{The circle should start by the diamond on the top platform} \label{fig:circlelevel01}
\end{figure}

\begin{figure}[h]
\centering
\begin{minipage}{.5\columnwidth}
\centering
\includegraphics[width=\columnwidth]{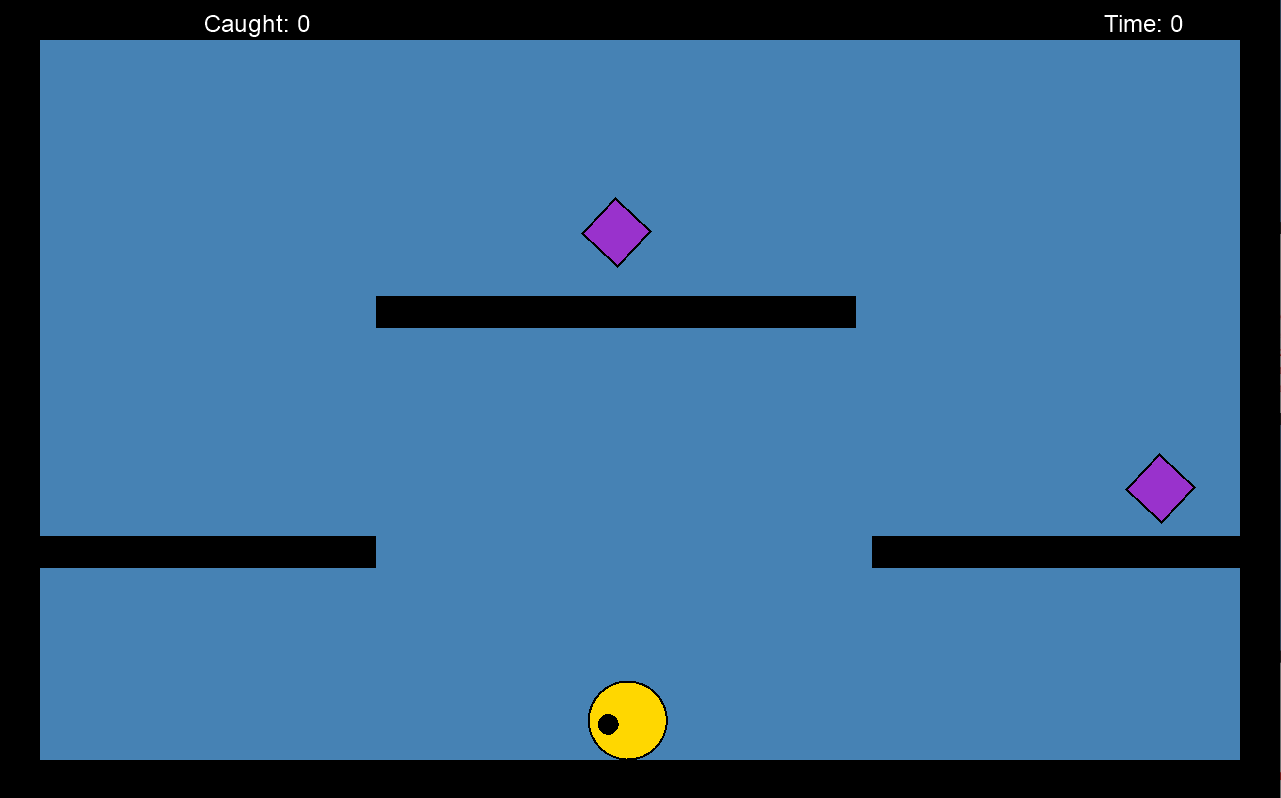}
\end{minipage}%
\begin{minipage}{.5\columnwidth}
\centering
\includegraphics[width=\columnwidth]{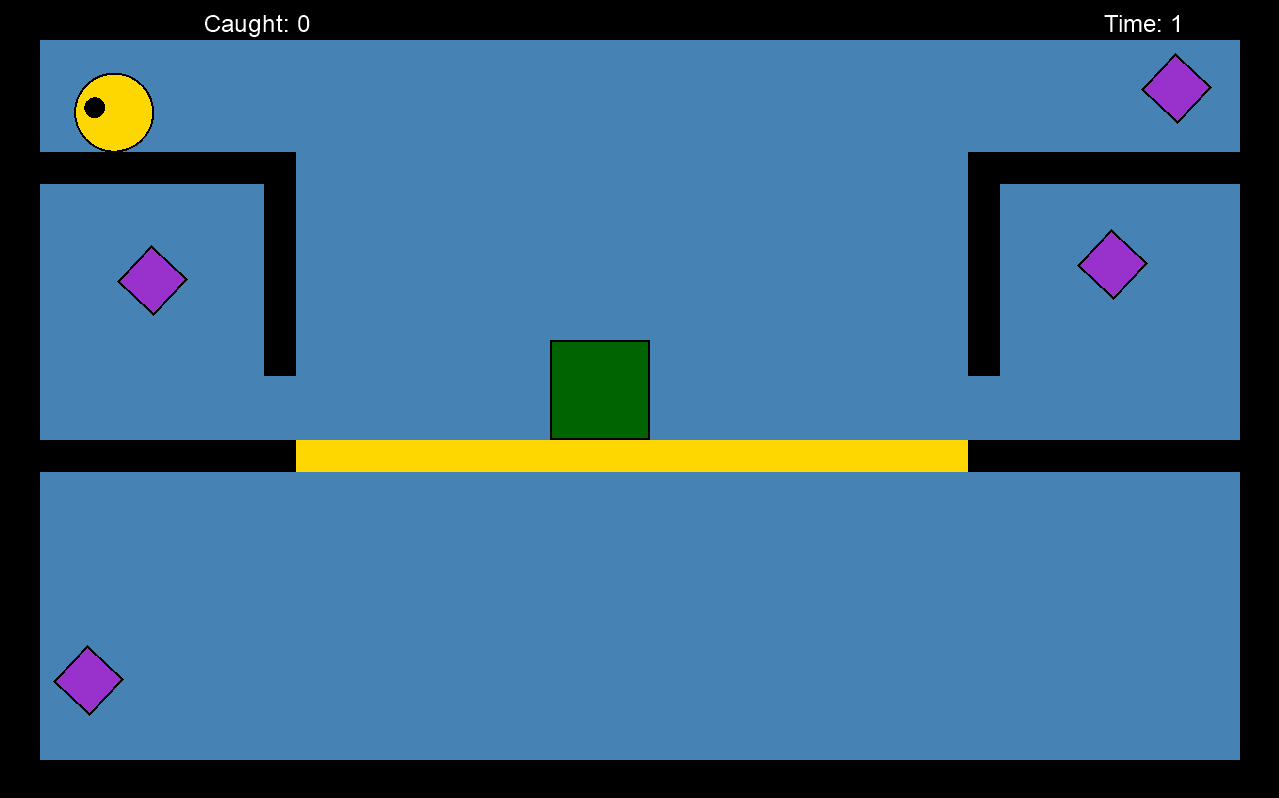}
\end{minipage}
\caption{Left: The circle should take the right platform first. Right: a level that combines three kinds of challenges} \label{fig:circlelevel03}
\end{figure}

So, in other words, task planning is about puzzle solving. Most of the levels of Geometry Friends present at least one type of puzzle to solve:
\begin{itemize}
    \item Deciding which diamond each character should catch, either by itself or with the help of the other. See Figure~\ref{fig:cooplevel04}.
    \item Finding the correct order to catch the diamonds. See Figure~\ref{fig:circlelevel01}.
    \item Finding a plan that reduces the path between the platforms. See Figure~\ref{fig:circlelevel03} (left).
    \item In general, it is a combination of the above. See Figure~\ref{fig:circlelevel03} (right).
\end{itemize}

Some levels may test effectiveness, by providing challenges where the order of the diamonds will influence the solvability of the level, like the one in Figure~\ref{fig:circlelevel01} (left). In this level, if the agent does not catch the diamond on the left first, then it will not be able to do it later. This serves to test if the agent is capable of plan carefully in situations with limited paths to take. On the other hand, all levels are tested for efficiency, through the time limit and the score which benefits the fastest agents. Figure~\ref{fig:circlelevel01} (right) and Figure~\ref{fig:circlelevel03} (left) are examples where the order does not matter to solvability but has impact on the time taken to finish the level.

When combining different types of puzzle, we are testing if the agent is able to deal with these problems in a balanced way. In Figure~\ref{fig:circlelevel03} (right) the circle should not catch the diamond on the bottom before catching the highest one on the right. To catch this, the circle needs the help of the rectangle. If the rectangle decides to catch the other two diamonds first, the circle will waste time waiting for the rectangle, while if they catch the highest one first, the circle can then go catch the one on the bottom while the rectangle catches the other two.

Motion planning deals with the timing of the low level actions that the agent can perform in the game. The approach for this kind of planning can be coupled more with the high level planing or with the motion control. But, in fact it is a bridge between the two. The motion planing is the process that makes sure that the agent performs correctly the actions that make the plan execution concrete in the game world.

\subsection{Motion Control}

Motion control is about executing the actions in the game world in a way that it follows the plan of action. In Geometry Friends, the characters' control is strongly coupled with the physics engine. For example, the movement is achieved by applying a force to the character (e.g. to move the circle a force is applied to make it roll). The collisions are realistic (as possible by the physics engine\footnote{We are using a Box2D implementation - https://box2d.org/.}). For example, the spin of the circle affects the collision. And attrition and gravity affect the movement as well. This means that predicting the movement of a character is not necessarily easy. Controlling a character in this context is closer to controlling a simulated vehicle than a common character in a platform game.

Moreover, the environment is dynamic, when playing cooperative levels, and non-deterministic in general. It is dynamic from a point of view of a character as the game state may change without the intervention of the character (e.g. it observes changes, but it did not perform any action). It is non-deterministic, mostly because of the physics engine. There are slight changes in the internal state of the physics engine that are not perceptible for the character, for this reason, in states that are perceived equally for the character the results of the same action may be different. This non-determinism is also due to the non-discrete nature of some of the control actions, that are of physics control nature. To move a character, the agent needs to turn on or off a force in the direction of the desired movement, but the effect depends on the current state of the character, for example, if it is moving to the opposite direction it does not move immediately to the desired direction, as it needs to decelerate beforehand. 

The above reasons make the motion execution, and planning, challenging. And task re-planning might be necessary as well.







\section{THE COMPETITION}\label{sec:Competition}
The Geometry Friends Game AI Competition\footnote{https://geometryfriends.gaips.inesc-id.pt/} has been running since 2013. The main objective of the competition is to present different cooperative challenges to \gls{AI} agents.

The competition includes two main tracks: the Cooperative Track and the Single AI Track that is divided in two sub-tracks: the Circle Track and the
Rectangle Track. Although the main track is the Cooperative Track, we also include single player tracks for participants that want to tackle the puzzle problem-solving and the characters’ control issues before undertaking the more demanding task of cooperation because, to excel in the cooperation task, agents will need good individual control and problem-solving capabilities.

Participants are free to use any approach and algorithm they believe will solve the levels, while baring in mind that the challenge is real-time. Levels have a time limit.




The ranking process ensures an unbiased comparison between all competitors. Submitted agents are evaluated and automatically tested through ten levels per track. Five of those levels are made public, to aid the agent development, while the other five are kept private until the end of the competition. The objective of keeping some levels private is to help understand if there is any type of overspecialization of the agents. If an agent solves all public levels, and none of the private ones, then the agent was specifically trained to solve the public ones, and may not be able to solve any other.

Each level has a time limit and, if the agents solution takes longer than this limit, the level is considered incomplete and the score is computed accordingly. Each submission is
evaluated in each level a total of ten times, to reduce the effects of chance due to the real-time execution of the physics engine.
This results in a total of 100 runs per submission. Agents are ranked by the total number of collectibles that they gather in each level as well as by the time they take to
complete the level. The final score of a level is the average score across the 10 runs of the level. The score of run $i$ is computed through the following formula:

\begin{equation}\label{Eq:Score}
\textsc{score}_{i} = V_{completed} \times \frac{ (T_{max} - t)} {T_{max}} + ( V_{collect} \times N_{collect} ),
\end{equation}
%


Where $V_{Completed}$ is a bonus awarded for completing the level, $T_{max}$ is the time limit, $t$ is the time the agent took to finish the game, $V_{Collect}$ is the score awarded per diamond collected and $N_{Collect}$ is the number of diamonds collected.

The final score is the sum of the scores of the ten levels of the respective track.

\section{THE FRAMEWORK}\label{sec:FrameWork}
The competition provides a framework to develop, run and test agents (e.g. it includes the executable of the game). Some examples of agents are provided and some starting guides as well.

\subsection{Agent development}\label{sec:AgentDevel}
The framework provide the baseline implementation for the two agents. This baseline contains the mandatory main classes that implement the agent's interfaces in C\# and project solution. The participants may organize the rest of the project as they see fit, as long as it is possible to compile the agents from another machine. 

The interface provides four input/output methods. Agents receives two types of sensor arrays as input. One at the beginning of the level, through the method $Setup()$, and the other at each update cycle, through the method $SensorsUpdate()$. 

The setup method shares the static information of the level and sends the following information through its arguments:

\begin{itemize}
    \item $nI$ - The total number of obstacles (platforms) and diamonds.
    \item $rI$ and $cI$ - The initial state of the rectangle and circle characters.
    \item $oI$, $rPI$ and $cPI$ - The information about the obstacles.
    \item $colI$ - The information about the diamonds.
    \item $area$ - Game area dimensions.
    \item $timeLimit$ - The time limit of the level.
\end{itemize}

Agents can also make their preparatory computations and loading of relevant data in the setup method.

The agents receive additional information through the sensors update method. The state is shared in the following arguments:

\begin{itemize}
    \item $nC$ - The number of uncaught diamonds.
    \item $rI$ and $cI$ - The current state of the rectangle and circle characters.
    \item $colI$ - The information of the diamonds in the level.
\end{itemize}

An obstacle, or platform, is represented by its position and size, and a diamond by its position. The circle character information contains the position, velocity and radius, and the rectangle character information contains the position, velocity and height.

At each update, the agent is expected to return an action within the possible ones, through the method $getAction()$. The agent can also perform computation in response to the $Update(TimeSpan time)$ method that is regularly called on each game update cycle.

The circle character may choose the following actions:

\begin{itemize}
    \item \textit{Jump.} This applies a force to give an upward impulse to the character. It only works if the circle is in solid ground. If this is chosen and the circle is moving in the air, the action has no effect.
    \item \textit{Roll left.} This applies a force to the left in the attempt to make the circle spin to the left and roll if it is on a platform. The action is always possible. The circle can spin on air. Applying a force to roll left does not make the circle to instantly start moving left. If the circle is spinning to the right, it is first necessary to counter the current rotation. While on, the angular velocity is changed until it reaches a maximum value.
    \item \textit{Roll right.} Similar to roll left, but the force is applied to spin the circle to the right. 
\end{itemize}

The rectangle character may choose the following actions:

\begin{itemize}
    \item \textit{Slide left.} This applies a force that attempts to push the Rectangle character to the left. As in the case of the roll action of the circle character, the movement may not start immediately to the left because of the character's current speed. While on, the velocity increases until it reaches its limit. 
    \item \textit{Slide right.} similar to Slide Left, but pushing the rectangle to the right;
    \item \textit{Morph up.} This reshapes the rectangle character, at a constant rate, increasing its height and reducing its width to keep a constant area (and mass). While on, the control is performed gradually until the height limit is reached. It can be performed while in the air.
    \item \textit{Morph down.} Similar to Morph Up but decreases the rectangle's height.

\end{itemize}

Both characters can choose not to perform an action or send a message to the other character. By sending a message they share and object that may contain any information. The message is delivered in the next game update and is handled by a specific method (i.e. $HandleAgentMessages(AgentMessage[] newMessages)$).

In this case no control is executed the characters will keep their current movement. Note that a stop action is not available. To stop the agent needs to perform a counter action (e.g. request the application of a force to the opposite direction of the current movement) or wait for the effects of friction. Friction has more effect on the rectangle, because it slides.

In addition, participants should be aware that characters may move without directly activating any movement controls, due to other moving objects (e.g. because of collisions between characters) in the game world. The constants used in the physics engine, such as, gravity, friction, and the force and torque values used in the characters’ controls are provided in the API.

\subsection{Additional Tools}
The framework provides additional tools that can be used to support the development of agents to the competition.


\subsubsection{Level Editor}
The Geometry Friends framework includes a level editor to allow the creation of new levels. This way, the developers can create all the scenarios they may need to train and test their agents. For example, agents that use machine learning algorithms may rely on a great number of different levels during their training phase. Creating simpler levels can also help developers to visually understand what isolated problems the agents are already (or not) capable of solving.

There are also packs available with levels that were used in previous competitions and in other contexts of the game (e.g. for human to human collaboration).

\subsubsection{Forward Model}
The framework includes a component to support a forward model. It clones the game state and is able to run a simulation (faster than the real-time pace of the game) resulting in a prediction of a future state, in a time interval, if a given action if performed.

\subsubsection{Batch Simulation}
It is possible to run the game in a batch. This means that developers can run their agents in a set of levels repeatedly with or without the game graphics on. If the graphics are off it is possible to run the game at different speeds. The tools write a report with the results of all individual executions of an agent per level. It is also possible to run different agents in the batch process, to support comparison, for example.
This batch simulation is, in fact, the core engine for running the competition and dealing automatically with the submissions received from the website.   

\subsubsection{Visual Debug}
While running the game, it is possible to turn on visual debug to get some additional visual information. This is drawn on top of the game's graphical user interface during the execution of the game. The developers can write their own information on this tool (e.g. lines, figures and text). There is a text debug output as well, but the visual feature is quite useful to help understanding the agents' behaviour at run-time. 

\section{CURRENT SOLUTIONS}\label{sec:solutions}
Along the years the competition has been running, several solutions have been proposed. 

In this section, we will briefly describe these solutions. These descriptions are mainly based on a report that is requested by the submission process. Submissions also include source-code that can be analysed to understand further their particular technical details. The source code can be found in the competition website \footnote{In http://gaips.inesc-id.pt/geometryfriends/, and in the new site https://geometryfriends.gaips.inesc-id.pt/, for submissions after 2018}. The best solutions have been kept as baseline agents for the competition.

Unfortunately, in the first editions of the competition the submission of source code was not mandatory, hence some descriptions in this paper lack some detail in the cases that the report was not sufficiently informative.

With this section, we want to understand how these agents deal with the following aspects:
\begin{itemize}
    \item The tracks the agents solve.
    \item Input processing.
    \item Planning algorithms.
    \item Motion control.
    \item Planning/Execution layer structure.
    \item Problems identified.
\end{itemize}
\subsection{Agents}
\subsubsection{\textbf{CIBot}}
The CIBot agents~\cite{CIBotCoop}\cite{CIBotRectangle}\cite{CIBotCircle}, have a different approach for each character, and for the cooperation track.

The agents make a first analysis of the level, checking were it is possible for them to move on, generating a directed graph. In the case of the rectangle, the graph includes transitions for 3 different shapes of the rectangle (full height, mid height and lowest height).

The rectangle agent uses a Monte-Carlo Tree Search (MCTS) algorithm. The directed graph extracted from the level analysis is converted into a tree to be used by the search. The initial graph and, consequently, the tree, has information on its edges about one of three shapes the rectangle needs to take to go from one node to the other. 

The circle agent uses the Dijkstra algorithm for path finding. The resulting path is then divided into an action plan, which is translated into the actions the agent needs to perform in order to reach every node of the plan. 

The control for each agent consists in a action queue that is filled by the planner. The control is simple and greedily moves the agent to the place represented by the graph node and there execute one of the actions: stay, jump or morph. 

For the cooperative levels, the agents first attempt to catch the diamonds they can by themselves, and then enter a ``cooperative state''. The authors define this state as having the circle above a morphed down rectangle, to which they call a ``riding position''. This allows the circle to reach higher diamonds or platforms. To get into this ``riding position'', the agents have a set of rules to follow, that involve placing the rectangle agent in the right position and wait for the circle to be on top on the rectangle, that then morphs up.

The authors identify some problems. The MCTS algorithm sometimes returns impossible paths, and the cooperation track does not deal with all cases. It failed in most private levels, despite its good performance on the public levels.

\subsubsection{\textbf{KUAS-IS Lab}}
The KUAS-IS Lab agents~\cite{KUAS-IS}, only play the single-player tracks. The two agents use a similar approach. They use A* search and Q-learning to solve the levels. A* search is used to find
the shortest path to go through all the diamonds in the level. This is based on a graph that represents
the level. The graph is similar to that of CIBot, but it also includes the diamonds’ positions. The
search heuristic uses the distance to the collectible; however, the agent tries to avoid situations that lead to pitfalls by just following a greedy approach, for example, if going to the closest diamond renders the level impossible. To address this problem, the search heuristic is weighted by the values of a Q-table that was built beforehand (offline) by training with Q-Learning. The control policy makes use of the Q-Learning offline training as well.

The authors recognize the hardships of controlling the characters, thus, suggest the use of a reinforcement learning approach.

\subsubsection{\textbf{OPU-SCOM 2014}}
The OPU-SCOM~\cite{OPU-SCOM} deals only with the rectangle agent for the single-player track. It uses and hierarchical approach dividing the agent in two layers.



The first layer searches for a global strategy while the second is related to the control of the character and searches for the sequence of actions to complete the strategy. In the first layer the level is converted into a graph by generating cells that cover the level. Cells are generated by tracing lines aligned with all platforms’ edges. Each cell becomes a node in the graph and neighbouring cells are connected. Nodes are added for the diamonds and for the initial position of the character. The latter are connected to the node that represents the cell they are in. Dijkstra’s algorithm is applied to define possible paths between the character and the diamonds, and a particle swarm optimisation algorithm (PSO) searches for the best order to pick-up the diamonds, given the possible paths. The agent selects the first diamond on the ordered list returned by the PSO and defines a hierarchical task plan composed by meta-tasks representing goals, such as, catching a diamond or falling down, and a set of sub-objectives. 

The second layer takes the meta-tasks, by the order defined in the plan, and selects and executes the actions needed to achieve it. It retrieves this information from a mapping between the meta-tasks to tasks adding the corresponding parameters. The agents uses four different tasks corresponding to the control actions: slide left, slide right, morph up and morph down. Tasks are executed until they reach successful termination that is checked every step (e.g. the character reached the targeted move position). Upon termination, the next task is started.

\subsubsection{\textbf{OPU-SCOM 2015}}

In 2015, the team submitted an updated version of the agent~\cite{OPU-SCOM2015}, that follows a similar hierarchical approach divided in two layers. The first one is used to compute the global strategy by first identifying (offline) the various sub-objective of the game and then generating the best strategy (sequence of sub-objectives) using a composition of genetic algorithm and neural network. The second layer then generates orders to perform the computed strategy.

The sub-objectives are defined in a list that was compiled through the identification of the relevant sub-objectives  by running the OPU-SCOM 2014 agent. The sub-objectives defined are related to changes in values of the properties of the game objects (e.g. increase or decrease the values of the coordinates X and Y, increase or decrease velocity in terms of X of Y, or increase or decrease the height of the rectangle). Then a strategy generator, based on a neural network using a Neuro Evolution of Augmenting Topologies (NEAT) approach, takes the game state, in run-time, as input and presents a single sub-objective as output. This sub-objective is then sent to the second layer. The second layer uses some hard coded rules that translate each sub-objective to its corresponding control action. For example, the sub-objective increasing X corresponds to the control action slide right.

\subsubsection{\textbf{RL Agent}}
The RL Agent \cite{RL} was developed to tackle the single-player circle track. But, it was never submitted to the competition. It was, nevertheless, the baseline for the development of the PG-RL Agent submitted to competition in 2015.

The RL Agent divided the problem of solving a Geometry Friends level in three sub-problems (SP): 
\begin{itemize}
    \item SP1 - Catching the diamonds on the platform the agent is currently on.
    \item SP2 - Deciding the next platform to go to.
    \item SP3 - Moving to a platform.
\end{itemize}

A Geometry Friends level is then solved by repeatedly
solving the series of (SP1 $\rightarrow$ SP2 $\rightarrow$ SP3) starting from solving the platform where the character is initially placed. 

The initial step, performed at setup time, is the creation of a navigational graph and the identification of platforms. A platform was defined as the region of an obstacle where the agent can move without having to jump. This means that the platforms defined in the level can be separated into more than one platform in the agent's representation of the game world. Then the diamonds in the level are assigned to platforms (to support SP1). They are assigned to the platform right below them.

SP1 and SP3 are treated as motion control problems and use a Reinforcement learning approach. When solving one platform (SP1) the character tries to catch all the diamonds assigned to the platform in a greedy way. The following features were used for learning: the position of the closest diamond on the platform, the circle position in relation to the edges of the platforms, the presence of a safe edge (blocked by an obstacle), the character's distance to the closest diamond, the velocity in X of the character and the number of diamonds in the platform. The reward was given by the number of collected diamonds, the number of diamonds to collect, the distance to the closest diamond and the time left until the level's time limit. SP3 uses additional features, such as, the distance between the platforms, the distance to the jump point on the edge and the size of the landing platform. The reward is the time taken to perform the jump. The training of the agent used a biased randomized action selection that make the agent move more often than jump during exploration.

To solve SP2, a \gls{DFS} search was used, taking from a starting point the agent's current position and trying to reach a platform that still has diamonds to collect. SP2 is triggered any time that agent is in a platform that has no diamonds to collect. Once SP2 finds a solution. SP3 is triggered to make the agent jump to the first platform returned by the search. If the agent is in a platform with diamonds to collect, SP1 is triggered.

The authors recognize that the assignment of the diamonds has some problems when diamonds are between platforms. Additionally, the greedy \gls{DFS} search will not perform well in complex puzzles. Some problems were found when trying to jump to small platforms.

\subsubsection{\textbf{PG-RL Agent}}
The PG-RL Agent \cite{PGRL} extended the RL Aproach described in the previous section to work for the rectangle and the cooperation tracks. It also improved the training set of the RL agent for the circle to include more situations for the agent to train, in particular to improve jumping from one platform to the another. The feature vector was simplified merging the information about position, size of the platform and speed of the character and adding symmetry to consider situations with symmetric coordinates to be treated as the same, hence, reducing the dimensionality of the game state. The sub-problem SP2 was solved by using the Dijkstra's algorithm instead of the \gls{DFS} to achieve shorter solutions that save time to improve the score.

The feature set for the rectangle was similar to the one used for the circle, only including and additional feature with the height of the character. The navigational graph was generated having into account the shape of the rectangle. The overall algorithm was the same, involving series of (SP1 $\rightarrow$ SP2 $\rightarrow$ SP3).

To address collaboration, a naive approach was used. The agent kept their initial training and trained further in a set of levels with cooperative challenges. A new feature, the position of the other character, was included. But it was only activated when the characters were close together. In the other cases the agents would represent the state as a single player problem. This makes it easier for the agent to reuse the previous knowledge. The planning part to solve SP2 did not change much. But, in the end the agent did not perform very well in the cooperation track.

\subsubsection{\textbf{RRT Agent}}
The RRT Agent \cite{GFRRT} uses the \gls{RRT} algorithm to address the single-player tracks.

This solution is divided in two layers, planning and control. The \gls{RRT} is used for planning, at the beginning of the level, which returns a path containing a sequence of what the authors define as ``points''. These points are states to reach, and can be descriptive or direct instructions. For example, one possible point is ``jump'' instructing the circle agent to jump, while another one is ``diamond above'' that indicates that a diamond is above the agent.

The controller then takes the information it needs from the path points, and computes an estimate of the velocity needed when reaching each point as well as determine when the agent should jump or morph. To help the agents achieve the correct velocity a \gls{PID} controller is used. Some of the points given by the plan already have motion instructions to be followed by the controller.

The authors recognize some problems related to the control of the characters, more specifically, of the rectangle.

\subsubsection{\textbf{RRT2017 Agent}}

The RRT2017 Agent~\cite{RRT-EPIA}, was developed using the RRT Agent as a baseline. It improved the state projections of the search, by using the forward model provided by the competition framework, that was not yet available for the first agent.

Using this forward model was more computationally expensive, but it was more precise. Some other extensions were included to the \gls{RRT} search to reduce the search time. A re-planning capability was also added, in case the planner was not able to find a complete solution at first, or if the agent fails to reach a desired state of the plan. Some of these extensions were included to deal with the cooperation track.

This agent uses a different controller than its baseline. This controller performs real-time motion planning by checking the current position and velocity of the agent and comparing it to the desired one. It then computes the action needed, depending on the acceleration it needs to achieve said velocity at said position. If the state of the plan has a jump action associated, then the agent jumps when it reaches said state.

The main problems pointed by the authors are related to the search time taken due to the use of the forward model. This is, in particular, a problem for the cooperative agents. The authors also note that the controller can fail in some tricky situations where the margins of error used to check the action results may prevent the success of the action.

\subsubsection{\textbf{Rule-based RRT Agent}}
The rule-based RRT agent~\cite{RB-RRT} picked up the RRT2017 agent as a baseline to improve the cooperation aspect of the agents, by identifying cooperation problems and creating a set of rules to deal with each. The agents have a single player mode and a cooperative mode. 

At the beginning of the level, each diamond is assigned to either agent, or both in case of requiring cooperation. In the latter case, the type of cooperation problem is also identified. The authors identify three possible cooperation problems:
\begin{itemize}
    \item Height: the circle cannot reach the diamond by itself, needing the rectangle to be used as a higher platform.
    \item Tight space: there is a diamond between two platforms only the rectangle can reach, but needs the help of the circle to climb the platform.
    \item Agent specific platforms: some platforms can only be used as such by one of the agents, the other needing the first to cross it.
\end{itemize}

The agents catch their assigned diamonds, first, and only then enter into cooperative mode, where they follow the rules according to the problem.

\subsubsection{\textbf{Subgoal A* Agent}}
The Subgoal A* Agent~\cite{SubGoalAStarThesis} uses a modification to the A* algorithm, the subgoal A*, to run the rectangle track.

The agent starts by creating an abstraction of the level to be used by the search algorithm, considering the abilities of the character. This results in a boolean matrix where the space occupied by obstacles is set as true. After that, a graph is created, where the edges indicate the action to be taken in order to move from one node to the other, including the shape the rectangle must take as well as the direction to follow.

The search is then performed, by the subgoal A*, an adaptation of the A* algorithm that adds a new property to a node. This property contains the diamonds that have been caught allowing the search to find a solution with all diamonds caught as a goal. 

The controller then executes the plan, using a set of 13 rules, according to the nodes of the plan.
\subsubsection{\textbf{KIT Agent}}
The KIT Agent~\cite{KIT} divides the problem into planning and its execution.

Similar to the RL Agent approach, there is a redefinition of the platforms to the regions where the agents can move on. From there, a directed graph is obtained by identifying from which platforms it is possible to reach other platforms and to collect diamonds. To understand this, trajectories are predicted by simulating the possible motion of the circle. The planning is done by applying the Subgoal A* algorithm, similarly to~\cite{SubGoalAStarThesis}, to the directed graph. 

The execution of the plan is done by the controller. This takes into account the positions and horizontal velocities given by the returned path. Then, like other previously mentioned agents, KIT Agent also selects the actions according to a previous Q-learning process.

The authors are aware of some limitations of this solution. For example, the agent may fail in levels with narrow platforms, that require small steps and jumps in order to be climbed, and levels with positions that can only be reached by colliding with platforms.
\subsubsection{\textbf{Supervised DL Agent}}
The Supervised DL Agent~\cite{Supervised} uses a deep learning approach for the circle character.

The agents developed their techniques by observing humans playing the game. They used the capture of screenshots and recorded the keys the human player pressed to perform the actions. This human-generated data was then used for the network training. 

During training, the action selection is $\epsilon$-greedy, after that, it is purely greedy.

The authors acknowledge that to follow successfully this kind of approach the agents need more data and training.

\subsubsection{\textbf{Neural Reinforcement Learning Agent}}
The Neural Reinforcement Learning Agent~\cite{NRL} uses a reinforcement learning approach that uses neural network function approximation to deal with the single-player circle track.

The network has an input layer, three hidden layers and a linear output layer. This output layer represents the approximate Q-value for each action available to perform at the input state. The network inputs consists on the distance to the obstacles around the agent in 32 different directions. 

This approach alternated between training, where the objective was to optimize the neural network to minimize the error of the value functions; and data gathering, where the agent interacted with the environment according to an $\epsilon$-greedy policy.

This author also recognizes that, such approach, needs more training time and data despite some improvements in the agent's neural architecture.
\subsubsection{\textbf{MARL-GF}}
The MARL-GF agent~\cite{MARLThesis} uses a \gls{MARL} approach for the cooperation track. The agents have three layers: trajectory analysis, planning and control.

The trajectory analysis layer takes into consideration the general motion equations for constant acceleration for both characters and also makes a climbing platform analysis in the case of the rectangle. For the cooperative levels, it also considers the riding position (when the circle stands on the rectangle and both characters move together).

The planning layer uses the Subgoal A* search on a directed graph, like some of the previously mentioned agents. For the cooperation levels, a directed graph is generated for each character and cooperative component and then merged into a single directed graph where the algorithm is performed

For the control, both characters use Q-Learning, as many others. As for the cooperation aspects, the document refers the use of either Team Q-Learning or Optimal Adaptive Learning for the training phase.
\subsubsection{\textbf{NKUST}}
The NKUST~\cite{NKUST} agents address cooperation as well.

The ``riding position'' is the main focus, as is it the most common type of collaboration through the presented levels. First, the area where the rectangle can move is computed, and, thus, which diamonds it can catch. With this, the circle character can get more information about the areas it can reach, since the rectangle is seen as a movable platform.

When searching for a path, the priority is catching the diamonds in an order that makes is possible to catch the highest number of diamonds by using a \gls{DFS} trial. Then an attempt to find a path is done with the A* algorithm. If no complete path is found, the agent follows the one with more diamonds it already discovered.

In terms of task division, the rectangle has priority over the diamonds. If this character can catch all of them by itself, then the circle is not used. Only when there is one or more diamonds that the rectangle cannot catch, the circle performs the path planning considering the motion space of the rectangle. Whenever the circle needs the rectangle to reach a position, this information is added to the ``collaborative nodes'' of the search. When the circle finishes its planning, then the rectangle performs another, with the information of the nodes that need collaboration if any exists.

To perform the cooperative tasks, an agent will wait for the other in case it reaches the start position for the cooperative motions before.
\subsubsection{\textbf{AGAgent}}
The AGAgent~\cite{AGAgent} performs several steps for planning for the circle track.

In the beginning of the level, the original state space is converted into abstract regions. These abstract regions are what before as been described as platforms, i.e. surfaces, referred as segments, where the character can move on. The agent then explores this new state space, using control policies, with the objective of identifying the possible transitions between the states. These policies consist on:
\begin{itemize}
    \item RollTo($x$, $vx$): Rolls the to the target $x$ position with $vx$ velocity.
    \item JumpAndStop: Jumps and forces the agent to stop when landing.
    \item FallAndStop: Lets the agent roll off the segment and forces it to stop when landing.
    \item RollUp: Rolls in attempt to climb a segment.
\end{itemize}
With this, a graph is generated and an A* search is performed. 

The return plan has already the actions necessary for its execution, as a result of the policies previously used during the initial exploration.

One of the problems the authors mention is related to the simulation phase where sometimes the predictions returned are not accurate.
\subsubsection{\textbf{Other Non-participant Agents}}
There are some agents developed for the Geometry Friends game that did not participate in the competition, but have been described in research publications.

Özgen et al.~\cite{Generalized} proposed a circle agent that also resorts to reinforcement learning. Similar to the Supervised DL Agent, the learning is done by having stacked frames as input via a convolutional neural network, where the resulting image is processed into states. Actions are then returned as output. The performance of the agent was compared with human and random agent performances. This approach was able to complete more games than the random agent, though it still had difficulties in doing such. The authors believe this problem may be due to the architecture of the network not being able to generalize all levels, or that it may need to be iterated with more frames so it can converge.

Simões et al.~\cite{Geo2} used asynchronous deep learning to create the agents' policies. The agents were tested with both \gls{A3C} and \gls{ANQ} approaches. The authors explained they initially had difficulties during training due either to the high cost of time per training steps or not enough amount of samples to achieve decent policies. In order to help speeding the training process, they used the breadcrumbs approach and input shaping techniques to decrease the complexity of the network

The authors of these agents implemented their own version of the game to develop their agents.

\subsection{Explored AI Problems}

In the following tables we can see a summary of the problems the agents deal with, and the algorithms/strategies they use. Table~\ref{tab:tracks} shows the tracks the agents try to solve. The planning algorithms and strategies used are summarized in table~\ref{tab:planning}. And table~\ref{tab:execution} shows the algorithms and strategies used for the plan execution/motion control.


\begin{figure}[h]
\centerline{\includegraphics[width=\columnwidth]{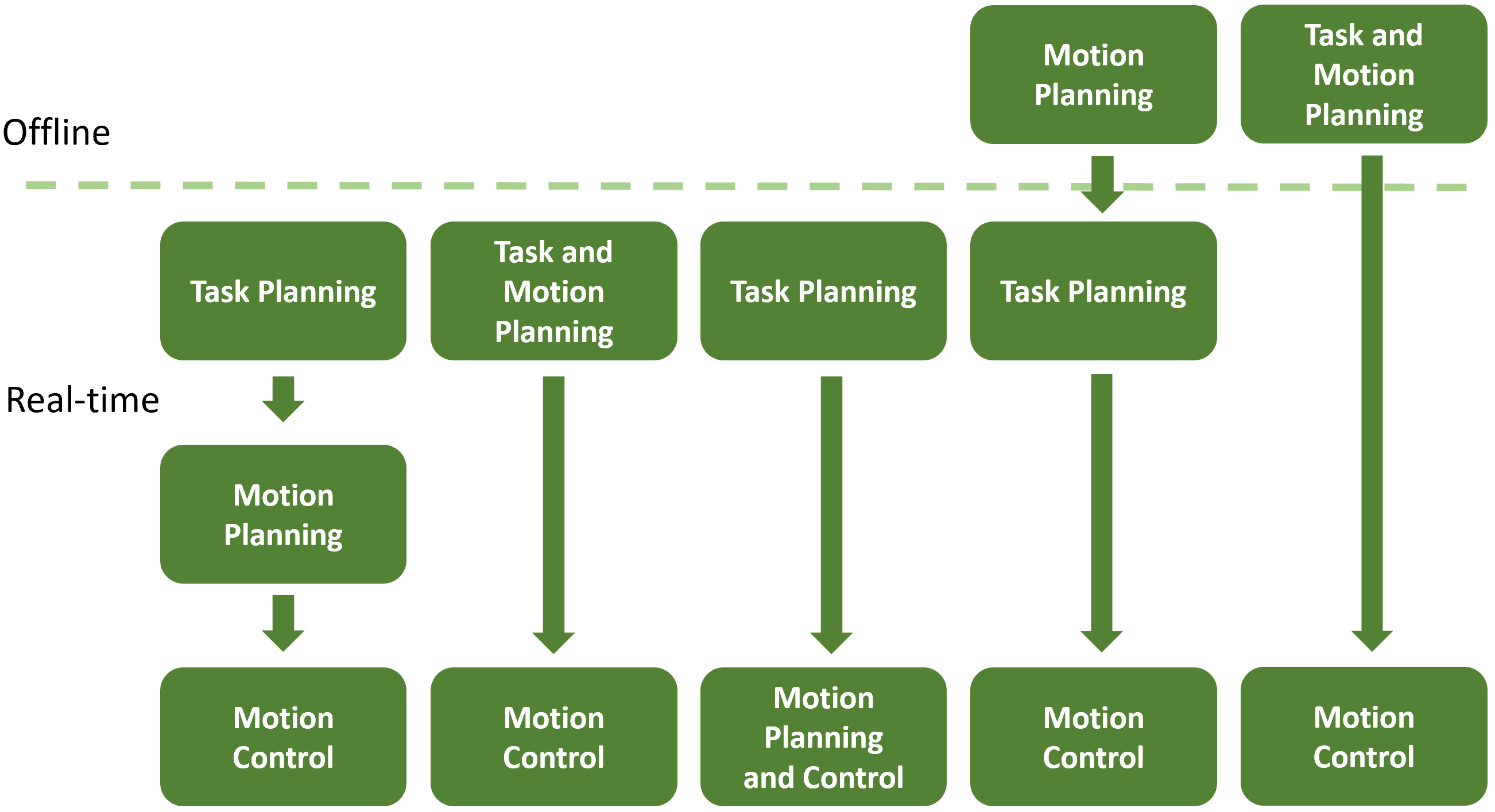}}
\caption{Examples of different layer separation for planning and control} \label{fig:planninglayers}
\end{figure}

Figure~\ref{fig:planninglayers} shows the different ways the agents described previously organise the layers of planning and execution. The first, on the left, has task planning, motion planning and planning execution in three separate layers. In this case, there is a first planner (task) that outputs a set of objectives that will be turned into a set of actions after going through the second planner (motion) to be then executed by the controller. In the second case both task and motion planning are done together. The difference here is that the task planning is done already considering the challenges of motion control. The third one, implies that the controller receives only the set of the planned objectives and, in real-time, for each objective, plans the necessary actions to follow. The fourth one represents the cases where the motion planning is done offline, usually through machine learning, or through the definition of rules, but the task planning is still done when the agent is running. In the last case, both task and motion planning are performed offline. In the table~\ref{tab:layers} we list the agents by the layer structure we believe they follow, according to their description. The numbers represent the structures from Fig.~\ref{fig:planninglayers} from left to right. It is possible to conclude that most agents use an offline motion planning approach.

The results of the 2019 competition, presented in table~\ref{tab:compResults}, show that MARL-GF and KIT agents are the most promising solutions at the moment. Still, they did not perform perfectly, having several levels of the tracks they were not able to finish, specially in the cooperation track.
\begin{table}[h]
\begin{tabular}{lccc}
\multirow{2}{*}{Agent} & \multicolumn{3}{c}{Tracks}                                                                   \\ \cline{2-4} 
                       & \multicolumn{1}{l}{Cooperation} & \multicolumn{1}{l}{Circle} & \multicolumn{1}{l}{Rectangle} \\ \hline
CIBot                  & Yes                             & Yes                        & Yes                           \\
KUAS-IS Lab            & No                              & Yes                        & Yes                           \\
OPU-SCOM 2014          & No                              & No                         & Yes                           \\
OPU-SCOM 2015          & No                              & No                         & Yes                           \\
RL Agent               & No                              & Yes                        & No                           \\
PG-RL                  & Yes                             & Yes                        & Yes                           \\
RRT                    & Yes                             & Yes                        & Yes                           \\
Subgoal A*             & No                              & No                         & Yes                           \\
RRT2017                & Yes                             & Yes                        & Yes                           \\
KIT                    & Yes                             & Yes                        & Yes                           \\
Supervised DL Agent    & No                              & Yes                        & No                            \\
NRL                    & No                              & Yes                        & No                            \\
MARL-GF                & Yes                             & Yes                        & Yes                           \\
NKUST                  & Yes                             & No                         & No                           \\
Rule-based RRT         & Yes                             & Yes                        & Yes                           \\
AGAgent                & No                              & Yes                        & No                           
\end{tabular}
\caption{\label{tab:tracks}Tracks agents are prepared for.}
\end{table}
\begin{table}[h]
\begin{tabular}{lccc}
\multirow{2}{*}{Agent} & \multicolumn{3}{c}{Planning}                                                                 \\ \cline{2-4} 
                       & \multicolumn{1}{l}{Cooperation} & \multicolumn{1}{l}{Circle} & \multicolumn{1}{l}{Rectangle} \\ \hline
CIBot                  & -                               & Dijkstra                   & MCTS                          \\
KUAS-IS Lab            & -                               & A*                         & A*                            \\
OPU-SCOM               & -                               & -                          & PSO                             \\
OPU-SCOM               & -                               & -                          & NEAT                             \\
RL Agent               & -                               & DFS                        & -                           \\
PG-RL                  & -                               & Dijkstra                        & Dijkstra                           \\
RRT                    & -                               & RRT                        & RRT                           \\
Subgoal A*             & -                               & -                          & Subgoal A*                    \\
RRT2017                & RRT                             & RRT                        & RRT                           \\
KIT                    & -                               & Subgoal A*                 & Subgoal A*                    \\
Supervised Agent       & -                               & DL              & -                             \\
NRL                    & -                               & RL     & -                             \\
MARL-GF                & MARL                            & Subgoal A*                 & Subgoal A*                    \\
NKUST                & A*                            & -                 & -                    \\
Rule Base RRT          & Rules                           & RRT                        & RRT                           \\
AGAgent                & -                               & A*                         & -                            
\end{tabular}
\caption{\label{tab:planning}Planning algorithms/strategies used by the agents.}
\end{table}
\begin{table}[h]
\begin{tabular}{lcc}
\multirow{2}{*}{Agent} & \multicolumn{2}{c}{Execution}                                                                                                                   \\ \cline{2-3} 
                       & Circle                                                                 & Rectangle                                                              \\ \hline
CIBot                  & Rule-based                                                                      & Rule-based                                                             \\
KUAS-IS Lab            & Q-Learning                                                             & Q-Learning                                                             \\
OPU-SCOM 2014               & -                                                                      & Rule-based                                                                      \\

OPU-SCOM 2015               & -                                                                      & Rule-based                                                                      \\
RL Agent              & RL                                                                     & -                                                        
             \\
PG-RL                  & RL                                                                     & RL                                                                     \\
RRT                    & Plan Execution + PID                                                   & Plan Execution + PID                                                   \\
Subgoal A*             & -                                                                      & Rule-based                                                             \\
RRT2017                & Motion planing                                                         & Motion planning                                                        \\
KIT                    & Q-Learning                                                             & Q-Learning                                                             \\
Supervised DL Agent       & DL                                                          & -                                                                      \\
NRL                    & RL                                                 & -                                                                      \\
MARL-GF                & Q-Learning                                                             & Q-Learning                                                             \\
NKUST                & Q-Learning                                                             & Q-Learning                                                             \\
Rule Base RRT          & \begin{tabular}[c]{@{}c@{}}Rule-based +\\ Motion Planning\end{tabular} & \begin{tabular}[c]{@{}c@{}}Rule-based +\\ Motion Planning\end{tabular} \\
AGAgent                & Plan Execution                                                         & -                                                                     
\end{tabular}
\caption{\label{tab:execution}Motion/Control algorithms used by the agents.}
\end{table}
\begin{table}[h]
\begin{tabular}{lllll}
\multicolumn{5}{c}{Structure}                                                                                                       \\ \hline
\multicolumn{1}{c}{1} & \multicolumn{1}{c}{2} & \multicolumn{1}{c}{3} & \multicolumn{1}{c}{4} & \multicolumn{1}{c}{5}               \\ \hline
OPU-SCOM 2014         & CIBot                & RRT2017               & KUAS-IS Lab           & Supervised                          \\
OPU-SCOM 2015         & RRT                   & RB RRT                & PG-RL                 & NRL                                 \\
                      & AGAgent               & NKUST                  & Subgoal A*            & \cite{Generalized} \\
                      &                       &                       & KIT                   & \cite{Geo2}        \\
                      &                       &                       & MARL-GF               &                                    
\end{tabular}
\caption{\label{tab:layers}Organisation of planning layers as presented in Figure \ref{fig:planninglayers}.}
\end{table}
\begin{table}[h]
\centering
\begin{tabular}{lll}
\multicolumn{1}{c}{Track}    & \multicolumn{1}{c}{Agent} & \multicolumn{1}{c}{\begin{tabular}[c]{@{}c@{}}Score\\ (approx.)\end{tabular}} \\ \hline
\multirow{4}{*}{Cooperation} & MARL-GF                   & 11887                                                                         \\
                             & RRT2017                   & 4490                                                                          \\
                             & Rule-based RRT             & 4166                                                                          \\
                             & NKUST                     & 3087                                                                          \\ \hline
\multirow{3}{*}{Circle}      & KIT                       & 14804                                                                         \\
                             & AGAgent                   & 14394                                                                         \\
                             & RRT2017                   & 8858                                                                          \\ \hline
Rectangle                    & MARL-GF                   & 13946                                                                         \\
                             & RRT2017                   & 10826                                                                         \\
                             & Subgoal A*                & 5042                                                                         
\end{tabular}
\caption{\label{tab:compResults}Results of the 2019 Competition}
\end{table}



\section{Conclusions and Future Work}\label{sec:conclusions}
Game AI competitions are important motivators for research and development on Game AI, which deals with issues that can be applied to more general AI problems.

We believe that the Geometry Friends Game AI Competition is a strong competition for it deals with several \gls{AI} problems, related to cooperation, task and motion planning, and control, all this with the need to work real-time. Some of these problems already explored by other competitions, but our competition is unique in the way it assembles them at the same time. 

The Geometry Friends is also a game that can be later expanded to had more challenges and deal with other AI problems.

First of all, the game has some features that have yet to be introduced, such as, moving platforms. These features are implemented in the game engine but left out of the competition levels, as we are currently waiting to get better results in the tracks as they are before adding more complexity to the base problem.

The competition may include, in the future, a level generation track as well, which we believe would be an interesting scenario for the development of PCG (Procedural Content Generation) algorithms. The creation of puzzle game scenarios is not something completely new in the PCG community, although we believe that the Geometry Friends game will prove an interesting test-bed for these algorithms. The main novelty is the cooperative nature of Geometry Friends, which adds some interesting challenges, as levels should to be fun for both players, for example. Additionally, making sure that the levels are solvable may actually be a hard problem. We already have a clear specification for the levels, which would facilitate the development of this track. Nevertheless, we would still require
an adequate framework to evaluate the levels generated. One
possibility would be to follow the current evaluation practices
in the level generation tracks where the ranking process consist of interleaving artificially generated levels with human-created levels and allow human players to play all. 

This may lead to the creation of a level generation tool to create unlimited number of levels automatically. This may be quite important for machine learning approaches that need large number of examples to be able to generalise the performance. And remove the bias of human designers as well, that may not include enough diversity in the levels.

It would also be interesting to explore a systematic way (eventually automatic) to categorize the levels in terms of the challenges they provide. This can guide the definition of specific sets of levels and support a better comparison of the agents being developed. In particular, it would be interesting to assess the difficulty of the levels.

Another extension that has been considered is an Agent
Believability track. This track would consist of creating agents
that would act in a human-believable way. The ranking process
could consist of a Turing-like test where users would view
various human and agent game-play sessions and try to identify which are human and which are artificial.

Finally, the last idea currently being developed, which is
in fact one of the main motivations for the development of the whole agent framework, is the Human-AI Cooperation track.
This will consist of developing a single AI agent (Circle or
Rectangle) that can play cooperatively with a human player.
Besides the challenges presented in the previous sections, this
track would require agents to effectively communicate with
human players, predict their movements and interact with their
characters in an entertaining way. The ranking of this track
would require tracking the performance of both players (such
as the number of levels solved and how long did it take to
solve them) and randomly have users play with other users
or agents (without their knowledge) and ask them how their
playing experience was. This will include similar concerns as
the believability track.

\section*{Acknowledgment}
This work was supported by national funds through FCT, Fundação para a Ciência e a Tecnologia, under project UIDB/50021/2020.
\bibliographystyle{IEEEtran}  
\bibliography{bib}  
\vspace{12pt}

\end{document}